\documentclass[runningheads]{llncs}
\usepackage[T1]{fontenc}
\usepackage{graphicx}
\usepackage{amsmath}
\usepackage{amssymb}     
\usepackage{booktabs}    
\usepackage{multirow}    
\usepackage{array}
\usepackage{xcolor}  
\usepackage{marvosym} 
\begin{document}
\title{Multi-scale Decomposed Convolution Refinement Network for Visible-Infrared Person Re-Identification}
\titlerunning{MDCRNet for Visible-Infrared Person Re-Identification}
%
%
\author{
Mingsheng Zheng\inst{1} \and
Zirui Jiang\inst{1} \and
Bo Liu\inst{1} \and
Yupeng Chen\inst{1} \and
Jun Zhang\inst{1} \and
Kai Zhao\inst{1,2}\textsuperscript{(\Letter)}
}

\authorrunning{M. Zheng et al.}

\institute{
School of Computer Science and Technology, Xinjiang University,
Urumqi, Xinjiang 830046, China
\and
Joint International Research Laboratory of Silk Road Multilingual
Cognitive Computing, Xinjiang University,
Urumqi, Xinjiang 830046, China\\
\email{zhawkk@xju.edu.cn}
}

%
\maketitle              
\begin{abstract}
Visible-infrared person re-identification (VI-ReID) suffers from cross-modal discrepancies and limited discriminative capabilities, leading to suboptimal recognition performance. Current approaches exhibit limitations in semantic mining, cross-modal fusion and feature constraints. To tackle these challenges, we propose MDCRNet, a Multi-scale Decomposed Convolution Refinement Network that enhances cross-modal feature learning and discriminative metric learning. Specifically, we introduce a Hierarchical Learning Module (HLM) containing four Hierarchical Decomposed Convolution Attention (HDCA) modules, each equipped with lightweight channel attention and multi-scale spatial perception blocks to capture multi-scale spatial dependencies. Moreover, we develop a Joint Discriminative Metric Loss (JDML) incorporating a novel Granularity Discriminative Loss (GDL) that simultaneously optimizes intra-identity compactness and inter-identity separability across modalities. Extensive experiments on SYSU-MM01 and RegDB datasets demonstrate that MDCRNet achieves state-of-the-art performance on both benchmarks. Code is available at https://github.com/Kevin-zms/MDCRNet.

\end{abstract}

\section{Introduction}
\label{sec:intro}

Person re-identification (ReID) aims to recognize individuals across different camera views, with critical applications in intelligent surveillance and public safety. Traditional ReID methods primarily focus on red-green-blue (RGB) images captured under visible light conditions, leading to severely degraded performance in nighttime, low-light, or adverse weather scenarios. To address these limitations, VI-ReID has emerged as a promising solution for continuous 24-hour surveillance by leveraging complementary information from both visible and infrared modalities. Nevertheless, VI-ReID encounters considerable difficulties stemming from pronounced cross-modal disparities between visible imagery that encompasses abundant chromatic information and textural details, and infrared imagery that records geometric boundaries and heat patterns, further complicated by within-modal variations such as perspective alterations, object obstructions, and apparel differences.

\begin{figure}[!ht]
\centering
\includegraphics[width=0.85\textwidth]{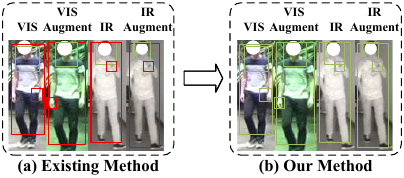}
\caption{Motivation of MDCRNet. (a) Existing methods adopt fixed-scale perception and capture only partial cues. (b) Our progressive multi-scale kernels jointly capture fine-grained (hands, watch, texture), mid-grained (face, clothing shape, upper-body pose), and coarse-grained (overall silhouette) information across the visible, infrared, and channel-augmented streams.}
\label{fig_motivation}
\end{figure}

Early VI-ReID approaches primarily relied on feature fusion strategies and semantic information learning techniques to bridge cross-modal discrepancies. With advances in deep learning, mainstream methods have evolved from traditional convolutional neural network (CNN) networks \cite{radenovic2018fine} to sophisticated architectures incorporating ResNet backbones, Transformer-based attention mechanisms \cite{chen2022structure}, and CLIP-assisted networks for enhanced cross-modal representation learning \cite{yu2025clip,ye2023channel}. Recent approaches \cite{feng2025learning,qiu2024high,feng2024cross,gao2021mso} focus on learning modality-invariant features through adversarial training or shared embedding spaces to minimize modal differences between visible and infrared modalities and designing effective loss functions to optimize models. However, existing methods exhibit significant limitations in three critical aspects. First, methods like MSCMNet \cite{hua2025mscmnet} employ efficient attention mechanisms to enhance cross-modal learning but rely on simple concatenation for fusion, overlooking inherent cross-modal data discrepancies. Second, channel-based approaches \cite{ye2021channel} partially address modality differences through channel augmentation yet neglect deep semantic mining and multi-scale spatial relationship learning. Third, existing loss functions typically address either inter-modal alignment or inter-identity discrimination independently, without providing unified constraints for diverse feature relationships including same-identity cross-modal, different-identity cross-modal, and intra-modal scenarios \cite{zheng2017discriminatively}. As shown in Fig. \ref{fig_motivation}(a), fixed-scale perception in existing methods captures only partial information and fails to jointly model fine-grained cues (hands, watch, texture), mid-grained cues (face, clothing shape, upper-body pose), and coarse-grained cues (overall silhouette). Our progressive multi-scale design (Fig. \ref{fig_motivation}(b)) captures all three granularities via kernels of increasing size, while channel attention adaptively fuses the cross-modal streams, jointly resolving the first two limitations. Furthermore, the resulting cross-modal cross-identity features are still poorly exploited by conventional losses during optimization, which motivates our unified discriminative metric to address the third limitation.

To address these limitations, we propose MDCRNet, a novel framework that effectively mitigates cross-modal discrepancies through hierarchical feature learning and discriminative optimization. Our approach begins with grayscale transformations \cite{ye2023channel} applied to both visible and infrared modalities, generating four complementary feature streams \cite{hua2025mscmnet} that reduce modal gaps while preserving essential structural information. Built upon a ResNet backbone, we introduce the HLM equipped with four HDCA modules, each containing Channel Attention (CA) for efficient cross-modal fusion and Multi-scale Spatial Perception Block (MSPB) for capturing hierarchical spatial dependencies. HDCA decomposes channels into groups with separate attention weights to disentangle modality-specific/shared features, while HLM is the sequential framework combining HDCA and Progressive Distance Perception Attention (PDPA). The MSPB incorporates three PDPA components with three kernel sizes, systematically extracting short-range, medium-range, and long-range spatial correlations for comprehensive semantic understanding. Furthermore, we design a JDML that integrates our novel GDL with existing identity and Quadruple Center Triplet (QCT) loss. The GDL employs three synergistic constraints: center loss to align same-identity cross-modal features, cosine similarity loss to separate different-identity features, and adaptive identity loss to enhance intra-modal clustering. Our three contributions are as follows:

\begin{itemize}

	\item We propose MDCRNet, an innovative architectural approach that effectively reduces cross-modal discrepancies through hierarchical decomposed convolution attention and progressive distance perception for VI-ReID.

	\item We design the HLM with HDCA modules that integrate CA for cross-modal fusion and multi-scale spatial perception blocks with PDPA, enabling extraction of discriminative semantic representations across different spatial ranges.

	\item We develop JDML incorporating a novel GDL with three complementary constraints that jointly optimize cross-modal alignment, inter-identity separation, and intra-modal consistency, demonstrating superior performance on SYSU-MM01 and RegDB benchmarks.
\end{itemize}

\section{Related Work}
\label{sec:related}

\subsection{Cross-Modal Person Re-Identification}
More broadly, recent advances in multimodal vision have emphasized efficient
and visually grounded instruction tuning \cite{dong2026ofa,dong2026visnec},
while extending visual understanding to synthetic long-form benchmarks and
memory-efficient streaming scenarios
\cite{dong2026objectstream,li2026pistachio}. Cross-modal person re-identification faces significant challenges in feature extraction and embedding alignment due to inherent modality discrepancies between visible and infrared images \cite{nguyen2017person}. Recent works have explored modality-shared networks to mitigate these discrepancies, yet often fail to capture multi-scale information such as texture, color, and structural cues \cite{ye2023channel,feng2025learning,ye2021channel}. Another line of research enhances semantic representation through attention mechanisms and modality conversion, but these approaches may inadvertently disrupt the original semantic structures \cite{zheng2024semi,wang2025learning,liu2021sfanet}. Furthermore, existing methods typically concatenate visible and infrared features, which proves insufficient for effective multimodal fusion \cite{hua2025mscmnet,chen2025hierarchical}. To address these challenges, we propose a multi-scale network that efficiently fuses and learns hierarchical information, providing a more robust framework for cross-modal representation learning. Unlike MSCMNet \cite{hua2025mscmnet}, which relies on simple feature concatenation and lacks unified multi-granularity constraints, and DSAF \cite{jiang2025dsaf}, whose dual-space alignment lacks fine-grained intra-class discriminative constraints, our HDCA and PDPA modules explicitly enforce both channel-wise modality disentanglement and distance-aware spatial hierarchy, providing the unified multi-granularity constraints that prior methods are missing.
   
\subsection{Modality Alignment}
Vision-language models have demonstrated remarkable progress across various vision tasks. Inspired by CLIP \cite{radford2021learning}, several VI-ReID frameworks have been proposed to leverage contrastive learning for cross-modal alignment. However, these approaches encounter challenges in effective feature alignment. For instance, SDCL \cite{yang2024shallow} integrates shallow and deep features via collaborative neighbor learning and ranking association, yet lacks a unified discriminative metric to jointly optimize cross-modality alignment and identity separation. Similarly, DSAF \cite{jiang2025dsaf} aligns visible-infrared modalities in dual spaces (Euclidean and Hilbert) with adaptive identity-consistent learning, yet lacks fine-grained discriminative constraints to simultaneously optimize intra-class compactness and inter-class separability. To address these limitations, we design a unified discriminative training objective that enforces cross-modal feature alignment while explicitly optimizing both intra-identity compactness and inter-identity separability.

\begin{figure}[!ht]
	\centering
	
	\includegraphics[width=\textwidth]{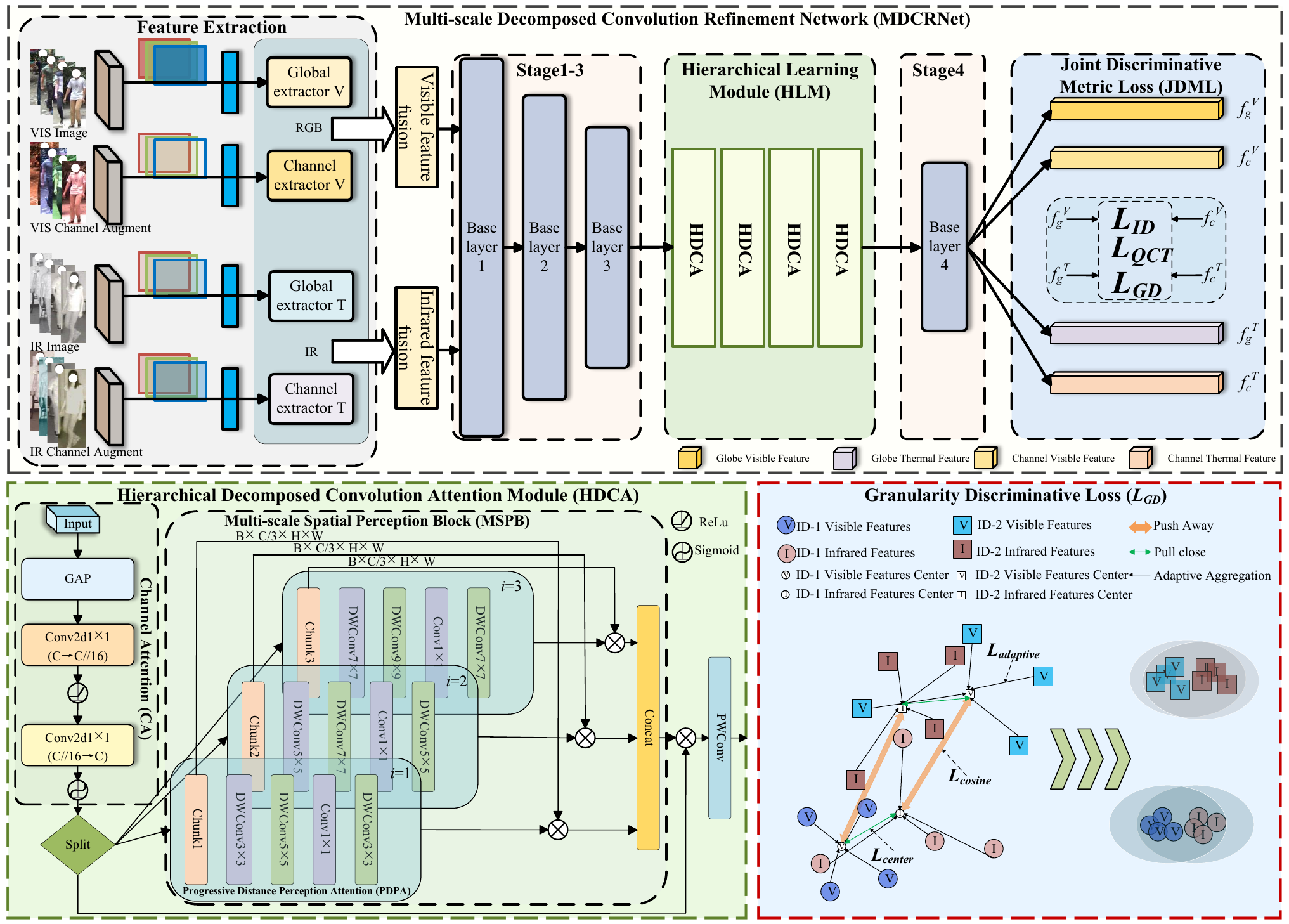}%
	\caption{MDCRNet employs a quad-stream architecture that processes cross-modal images along with their channel-augmented variants, then utilizes ResNet backbone for hierarchical feature extraction. The HLM integrates four HDCA modules, each containing CA and MSPB with three PDPA components to capture multi-scale spatial dependencies for effective cross-modal semantic learning. The JDML combines identity loss, QCT loss, and our proposed GDL.}
	\label{fig_sim}
\end{figure}
\section{Method}
\label{sec:method}

In visible-infrared person re-identification, given training samples $\{\mathbf{x}^v_i, \mathbf{x}^t_i, y_i\}_{i=1}^{N}$ where $\mathbf{x}^v_i \in \mathbb{R}^{H \times W \times 3}$ and $\mathbf{x}^t_i \in \mathbb{R}^{H \times W \times 3}$ denote visible and infrared images of identity $y_i$, the goal is to learn discriminative embeddings $\mathbf{f}^v_i, \mathbf{f}^t_i \in \mathbb{R}^{D}$ that minimize intra-identity cross-modal distances while maximizing inter-identity separability. In this section, we present MDCRNet to effectively mitigate cross-modal discrepancies while enhancing discriminative feature learning for VI-ReID, as illustrated in Fig. \ref{fig_sim}. Our approach adopts a quad-stream architecture that simultaneously processes original visible and infrared images alongside their channel-augmented variants, generating complementary feature representations through separate extraction pathways. The framework achieves effective cross-modal learning through two key innovations, namely the HLM equipped with four strategically positioned HDCA modules that capture multi-scale spatial dependencies through progressive distance perception mechanisms, and the JDML that jointly optimizes feature discrimination across modalities. During training, the HLM enhances cross-modal semantic alignment through efficient CA and multi-scale spatial perception, while JDML enforces fine-grained constraints on intra-identity compactness and inter-identity separability to achieve superior recognition performance.

\subsection{Hierarchical Learning Module}

The core challenge in VI-ReID lies in extracting discriminative features that are both modality-invariant and identity-specific. Existing methods often struggle to effectively fuse cross-modal information while capturing multi-scale spatial dependencies for comprehensive semantic understanding. To overcome this constraint, we introduce the HLM, which employs HDCA to facilitate efficient cross-modal learning and extract fine-grained discriminative features across different spatial ranges. The HLM contains four strategically positioned HDCA modules, where each HDCA integrates a CA mechanism and a MSPB. Each MSPB incorporates three PDPA components with kernel sizes 3, 5, and 7 to capture short-range, medium-range, and long-range spatial correlations respectively.

\subsubsection{Efficient Channel Attention.}

To effectively integrate the extracted cross-modal features and enhance the cross-modal perception capabilities of the subsequent MSPB, we propose a lightweight channel gating attention mechanism that efficiently fuses feature representations from four distinct data streams:

\begin{equation}
	\mathbf{G} = \sigma(\text{Conv}_{1 \times 1}(\text{ReLU}(\text{Conv}_{1 \times 1}(\text{GAP}(\mathbf{X}_{comp})))))
\end{equation}

\noindent where $\text{GAP}$ denotes Global Average Pooling, $\sigma$ is the Sigmoid activation and $\mathbf{X}_{comp} \in \mathbb{R}^{B \times C \times H \times W}$ denotes the concatenated feature representation from the four input streams.

\subsubsection{Progressive Distance Perception Attention.}

We employ three parallel PDPA modules that extract feature representations at different granularity levels:

\textbf{Three parallel PDPA ($\text{PDPA}_{3,5,7}$):} These modules, each with varying convolutional depths, facilitate the extraction of high-order structural information and semantic associations, particularly enabling the recovery of modality-specific semantic information that may be attenuated in shallow network layers:

\begin{align}
    \label{eq2}
	\text{PDPA}_{2i+1}(\mathbf{X}_i) &= \text{Conv}_{1 \times 1}(\text{DWConv}_{(2i+3) \times (2i+3)}^{d=i+1} \notag \\
	&\quad (\text{DWConv}_{(2i+1) \times (2i+1)}(\mathbf{X}_i)))
\end{align}

\noindent where $\text{DWConv}$ denotes depthwise convolution, $i$ is the PDPA branch index ($i = 1, 2, 3$), and $d = i + 1$ represents the dilation rate that progressively expands the receptive field. Compared with ASPP-style designs \cite{chen2017deeplab} that use fixed-size kernels with independent dilations (effective receptive field $\text{ERF} = 2r+1$), our PDPA employs cascaded progressive kernels with progressive dilations, achieving quadratic growth $\text{ERF} = 2i^2 + 6i + 3$. The receptive field thus grows quadratically rather than linearly, enabling exponential expansion of spatial context while maintaining efficient parameter usage through depthwise separable convolutions. This accelerated expansion enables hierarchical distance-aware feature extraction from local to global contexts, directly addressing VI-ReID's core challenge of discovering modality-invariant structural patterns across spatial scales while suppressing modality-specific appearance.

\subsection{Joint Discriminative Metric Loss}

To optimize the overall model performance, we propose the JDML which encompasses Identity loss \cite{zheng2017discriminatively}, QCT loss \cite{hua2025mscmnet} and our novel GDL. The GDL simultaneously enforces intra-identity compactness and inter-identity separability across different modalities through fine-grained constraints. Let $\mathbf{f}^v_i$
 and $\mathbf{f}^t_i$ denote normalized feature representations of the $i$-th identity from visible and thermal modalities, respectively. The GDL operates through three synergistic constraints to optimize diverse feature relationships across modalities and identities.
\subsubsection{Center Loss for Cross-Modal Alignment.}

To minimize the spatial separation among identical-identity feature representations across distinct modal domains, we employ a center loss that pulls cross-modal features toward unified identity centers:

\begin{equation}
	\mathcal{L}_{center} = \frac{1}{N} \sum_{i=1}^{N} \left( \|\mathbf{f}^v_i - \mathbf{c}_i\|_2^2 + \|\mathbf{f}^t_i - \mathbf{c}_i\|_2^2 \right)
\end{equation}

\noindent where $\mathbf{f}^v_i, \mathbf{f}^t_i \in \mathbb{R}^D$ denote normalized $D$-dimensional features from visible and thermal modalities for the $i$-th identity, $\mathbf{c}_i = \frac{1}{2}(\mathbf{f}^v_i + \mathbf{f}^t_i)$ represents the unified center for the $i$-th identity, and $N$ is the number of identities in a batch.

\subsubsection{Cosine Similarity Loss for Inter-Identity Separation.}

To enhance inter-identity discriminability while maintaining intra-identity consistency across modalities, we introduce a bidirectional cosine similarity loss that operates on cross-modal identity relationships:

\begin{equation}
	\mathcal{L}_{cosine} = \mathcal{L}_{v \rightarrow t} + \mathcal{L}_{t \rightarrow v}
\end{equation}

\noindent where:

\begin{equation}
	\mathcal{L}_{v \rightarrow t} = \frac{1}{B} \sum_{i=1}^{B} \max(0, \max_{j \neq i} S_{ij} - S_{ii} + m)
\end{equation}

\begin{equation}
	\mathcal{L}_{t \rightarrow v} = \frac{1}{B} \sum_{i=1}^{B} \max(0, \max_{j \neq i} S_{ji} - S_{ii} + m)
\end{equation}

\noindent Here, $\mathbf{S} = \mathbf{F}^v(\mathbf{F}^t)^T$ represents the cosine similarity matrix between visible and thermal features, where $B$ is the batch size. $S_{ii}$ represents the similarity between the $i$-th visible feature and its corresponding thermal feature (same identity), while $\max_{j \neq i} S_{ij}$ represents the maximum similarity between the $i$-th visible feature and thermal features from different identities.

\subsubsection{Adaptive Feature Aggregation for Identity Consistency.}

We propose an adaptive aggregation mechanism. For each identity group containing $K=4$ samples, we compute adaptive weights based on inter-sample similarities. Here $K=4$ follows the standard $P\times K$ sampling strategy used in MSCMNet \cite{hua2025mscmnet} and AGW \cite{ye2021deep}, where each batch contains $P$ identities with $K=4$ instances per modality. We empirically find that $K=2$ degenerates to a softmax over a single pair, while $K\geq 6$ inflates memory cost without further gain; $K=4$ thus balances discriminability and computational cost. The adaptive weights are computed as:
\begin{equation}
    s_i = \sum_{j=1}^{K} \text{sim}(\mathbf{f}_i, \mathbf{f}_j), \quad i = 1, \ldots, K
\end{equation}
\begin{equation}
    \mathbf{w} = \text{softmax}([s_1, s_2, s_3, s_4])
\end{equation}
\noindent where $\text{sim}(\cdot, \cdot)$ denotes cosine similarity. The aggregated identity representation is then computed as:
\begin{equation}
    \mathbf{f}_{agg} = \sum_{i=1}^{K} w_i \mathbf{f}_i
\end{equation}
\noindent where $w_i$ represents the adaptive weight for the $i$-th sample within the identity group.

The adaptive identity loss enforces compactness between adaptively aggregated features from the same identity across different modalities:
\begin{equation}
    L_{adaptive} = \frac{1}{B} \sum_{i=1}^{B} \|\mathbf{f}_{agg,i}^v - \mathbf{f}_{agg,i}^t\|_2
\end{equation}
\noindent where $\mathbf{f}_{agg,i}^v$ and $\mathbf{f}_{agg,i}^t$ are the adaptively aggregated features for the $i$-th identity from visible and thermal modalities respectively, and $B$ is the number of identities in a batch.

\subsubsection{Overall Loss Function.}

The complete Granularity Discriminative Loss combines all three components with appropriate weighting:

\begin{equation}
    \label{eq9}
	\mathcal{L}_{GD} = \lambda_1\mathcal{L}_{center} +\lambda_2\mathcal{L}_{cosine} +  \lambda_3\mathcal{L}_{adaptive}
\end{equation}

\noindent where $\lambda_1,\lambda_2,\lambda_3$ are set to 0.1. Note that $\mathcal{L}_{ID}$ enforces discriminative feature learning via softmax but lacks explicit cross-modal alignment; $\mathcal{L}_{QCT}$ pulls samples toward class centers using Euclidean distance but ignores angular information and direct cross-modal constraints. Our GDL addresses these limitations through three complementary components: $\mathcal{L}_{center}$ explicitly aligns cross-modal features and compensates for $\mathcal{L}_{ID}$'s absence of cross-modal supervision, $\mathcal{L}_{cosine}$ introduces angular metrics beyond Euclidean distance, and $\mathcal{L}_{adaptive}$ provides instance-level adaptive weighting beyond $\mathcal{L}_{QCT}$'s fixed centers.

The total training objective combines the  Identity loss, QCT loss, and GDL:

\begin{equation}
	\mathcal{L}_{JDM} = (\mathcal{L}_{ID} + \mathcal{L}_{QCT}) + \mathcal{L}_{GD}
\end{equation}

\section{Experiments}
\label{sec:experiments}

\subsection{Experimental Setup}

\noindent\textbf{Datasets.} We perform comprehensive experimental evaluations using two well-established VI-ReID benchmark datasets: SYSU-MM01 \cite{wu2017rgb} and RegDB \cite{nguyen2017person}.

\noindent\textbf{Evaluation Metrics.} In accordance with established research methodologies \cite{wang2019rgb,lu2023tri}, we utilize Cumulative Matching Characteristics (CMC) and mean Average Precision (mAP) as assessment criteria for our experimental evaluation.

\noindent\textbf{Implementation Details.} MDCRNet architecture is developed utilizing PyTorch framework and deployed on an NVIDIA A40 GPU platform, employing ResNet-50 as the foundational network structure. The HDCA module incorporates CA and three PDPA components with kernel sizes 3, 5, 7. The training process utilizes stochastic gradient descent (SGD) optimizer configured with a momentum parameter of 0.9 and regularization through weight decay set to $5\times10^{-4}$. The initial learning rate starts at 0.001 and increases to a maximum of 0.01, then undergoes reduction by a multiplicative factor of 0.1 at training epochs 30, 90, and 120 throughout a complete training duration of 220 epochs. Three components of GDL are weighted with 0.1.

\noindent\textbf{Complexity Analysis.} MDCRNet has 41.8M parameters (of which HDCA contributes 8.1M) and 20G FLOPs (HDCA contributes 1.7G). Inference speed is approximately 10ms per image on an NVIDIA A40 GPU, and end-to-end training takes about 20 hours on SYSU-MM01. These figures are comparable to recent VI-ReID baselines such as AGW and MSCMNet, confirming that the accuracy gains do not come at the cost of significant computational overhead.

\subsection{Comparison with State-of-the-Art Methods}

Our proposed MDCRNet is evaluated against contemporary state-of-the-art (SOTA) VI-ReID approaches using both SYSU-MM01 and RegDB benchmark datasets. Table \ref{tab1} and Table \ref{tab2} show the optimal performance obtained by the developed MDCRNet, which surpasses the second-best method on SYSU-MM01 dataset by 0.75\% and 0.61\% in all-search and indoor-search modes respectively, and on RegDB dataset by 1.06\% and 0.44\% in V-I and I-V modes, respectively. Notably, MDCRNet outperforms both Transformer-based methods (SPOT \cite{chen2022structure}, PMT \cite{lu2023learning}) and CLIP-assisted approaches (CAJ \cite{ye2023channel}), validating that our hierarchical decomposed convolution design, despite being CNN-based, effectively captures cross-modal cues that more complex architectures may overlook.

\begin{table}[!ht]
\centering
\caption{Comparison with SOTA methods on SYSU-MM01. R-$k$ recognition accuracy (\%) and mAP (\%) are presented.}
\label{tab1}
\setlength{\tabcolsep}{4pt}
\begin{tabular}{l|c|ccc|ccc}
\toprule
\multirow{2}{*}{Method} & \multirow{2}{*}{Publish} & \multicolumn{3}{c|}{All Search} & \multicolumn{3}{c}{Indoor Search} \\
 &  & R-1 & R-10 & mAP & R-1 & R-10 & mAP \\
\midrule
AGW \cite{ye2021deep} & TPAMI 21 & 47.50 & 84.39 & 47.65 & 54.17 & 91.14 & 62.97 \\
MPANet \cite{wu2021discover} & CVPR 21 & 70.58 & 96.21 & 68.24 & 76.74 & 98.21 & 80.95 \\
SPOT \cite{chen2022structure} & TIP 22 & 65.34 & 92.73 & 62.25 & 69.42 & 96.22 & 74.63 \\
CAL \cite{wu2023learning} & ICCV 23 & 42.40 & 85.00 & 40.70 & 45.90 & 87.60 & 54.30 \\
PMT \cite{lu2023learning} & AAAI 23 & 67.53 & 95.36 & 51.86 & 71.66 & 96.73 & 76.52 \\
DEEN \cite{zhang2023diverse} & CVPR 23 & 74.70 & 97.60 & 71.80 & 80.30 & 99.00 & 83.30 \\
MUN \cite{yu2023modality} & ICCV 23 & 76.24 & 97.84 & 73.81 & 79.42 & 98.09 & 82.06 \\
DMA \cite{cui2024dma} & TIFS 24 & 74.57 & - & 70.41 & 82.85 & - & 85.10 \\
MUCG \cite{zheng2024semi} & MM 24 & 68.80 & - & 65.90 & 77.40 & - & 81.00 \\
CAJ \cite{ye2023channel} & TPAMI 23 & 71.48 & 96.23 & 68.15 & 78.36 & 98.36 & 81.98 \\
MSCMNet \cite{hua2025mscmnet} & PR 25 & \textbf{78.53} & 97.51 & 74.20 & 83.00 & 98.99 & 85.54 \\
DSAF \cite{jiang2025dsaf} & TMM 25 & 76.65 & 97.74 & 73.24 & 83.48 & \textbf{99.27} & 83.78 \\
FMCNet+ \cite{xi2024fmcnet} & TNNLS 25 & 75.34 & - & 71.14 & 77.01 & - & 84.25 \\
HTCR \cite{chen2025hierarchical} & TMM 25 & 74.60 & - & 72.11 & 82.99 & - & 85.63 \\
\midrule
\textbf{MDCRNet} & \textbf{Ours} & 78.38 & \textbf{97.88} & \textbf{74.95} & \textbf{83.76} & 98.76 & \textbf{86.24} \\
\bottomrule
\end{tabular}
\end{table}

\begin{table}[!ht]
\centering
\caption{Comparison with SOTA methods on RegDB. R-$k$ recognition accuracy (\%) and mAP (\%) are presented.}
\label{tab2}
\setlength{\tabcolsep}{6pt}
\begin{tabular}{@{}l|c|cc|cc@{}}
\toprule
\multirow{2}{*}{Method} & \multirow{2}{*}{Publish} & \multicolumn{2}{c|}{V--I} & \multicolumn{2}{c}{I--V} \\
 &  & R-1 & mAP & R-1 & mAP \\
\midrule
AGW \cite{ye2021deep} & TPAMI 21 & 70.00 & 66.30 & 70.40 & 65.90 \\
SPOT \cite{chen2022structure} & TIP 22 & 80.30 & 72.40 & 79.30 & 72.20 \\
PMT \cite{lu2023learning} & AAAI 23 & 84.81 & 76.55 & 84.16 & 75.13 \\
SFANet \cite{liu2021sfanet} & TNNLS 21 & 76.31 & 68.00 & 70.20 & 63.80 \\
LCNL \cite{yang2024robust} & IJCV 24 & 85.60 & 60.60 & 84.00 & 76.90 \\
SDCL \cite{yang2024shallow} & CVPR 24 & 86.91 & 78.92 & 85.76 & 77.25 \\
MUCG \cite{zheng2024semi} & MM 24 & 86.90 & 76.70 & 83.70 & 74.10 \\
MSFCS \cite{yang2024cooperative} & TMM 24 & 85.34 & 76.39 & 83.88 & 75.16 \\
CAJ \cite{ye2023channel} & TPAMI 23 & 85.69 & 79.70 & 84.88 & 77.80 \\
CNGC \cite{wang2025learning} & ICASSP 25 & 86.31 & 79.90 & - & - \\
CM2GT \cite{feng2025learning} & PR 25 & 86.72 & 77.79 & \textbf{86.47} & 77.51 \\
\midrule
\textbf{MDCRNet} & \textbf{Ours} & \textbf{88.71} & \textbf{80.96} & 86.10 & \textbf{78.24} \\
\bottomrule
\end{tabular}
\end{table}

\begin{table}[!ht]
\centering
\caption{Ablation study of HDCA and GDL on SYSU-MM01. R-$k$ accuracy (\%) and mAP (\%) are presented.}
\label{tab3}
\setlength{\tabcolsep}{5pt}
\begin{tabular}{cc|cc|ccc|ccc}
\toprule
\multicolumn{2}{c|}{HDCA} & \multicolumn{2}{c|}{$\mathcal{L}_{JDM}$} & \multicolumn{3}{c|}{All Search} & \multicolumn{3}{c}{Indoor Search} \\
CA & PDPA & $(\mathcal{L}_{ID}+\mathcal{L}_{QCT})$ & $\mathcal{L}_{GD}$ & R-1 & R-10 & mAP & R-1 & R-10 & mAP \\
\midrule
 &  & \checkmark &  & 76.29 & 97.20 & 71.84 & 81.55 & 98.27 & 84.35 \\
 & \checkmark & \checkmark &  & 77.28 & 97.03 & 73.24 & 82.21 & 98.73 & 84.95 \\
\checkmark & \checkmark & \checkmark &  & 76.60 & 97.56 & 73.47 & 83.51 & 98.28 & 85.95 \\
\midrule
 &  & \checkmark & \checkmark & 74.86 & 97.02 & 71.41 & 81.20 & 98.76 & 84.47 \\
 & \checkmark & \checkmark & \checkmark & 77.16 & 97.48 & 73.54 & 82.62 & 98.42 & 85.34 \\
\checkmark & \checkmark & \checkmark & \checkmark & \textbf{78.38} & \textbf{97.88} & \textbf{74.95} & \textbf{83.76} & \textbf{98.76} & \textbf{86.24} \\
\bottomrule
\end{tabular}
\end{table}

\begin{table}[!ht]
\centering
\caption{Leave-one-out ablation of GDL components on SYSU-MM01 (all-search).}
\label{tab:gdl_loo}
\setlength{\tabcolsep}{10pt}
\begin{tabular}{lc}
\toprule
Configuration & mAP (\%) \\
\midrule
Full GDL & \textbf{74.95} \\
w/o $\mathcal{L}_{adaptive}$ & 73.27 \\
w/o $\mathcal{L}_{center}$ & 73.64 \\
w/o $\mathcal{L}_{cosine}$ & 73.77 \\
\bottomrule
\end{tabular}
\end{table}

\subsection{Ablation Study}

\noindent\textbf{Effectiveness of Key Components.} We conduct comprehensive ablation studies with different component combinations on SYSU-MM01 dataset. As shown in Table \ref{tab3}, the PDPA with multi-scale kernels effectively captures multi-scale spatial dependencies, improving R-1 accuracy from 74.86\% to 77.16\% in all-search mode. The CA mechanism within HDCA further enhances cross-modal feature alignment, achieving 83.76\% R-1 accuracy and 86.24\% mAP in indoor mode. The introduction of our $\mathcal{L}_{GD}$ significantly boosts performance by jointly optimizing $\mathcal{L}_{center}$, $\mathcal{L}_{cosine}$, and $\mathcal{L}_{adaptive}$, reaching the best mAP of 74.95\% in all-search mode. To further validate the contribution of each GDL component, we conduct leave-one-out ablations on SYSU-MM01 (all-search). As shown in Table \ref{tab:gdl_loo}, removing $\mathcal{L}_{adaptive}$, $\mathcal{L}_{center}$, or $\mathcal{L}_{cosine}$ individually drops the mAP from 74.95\% to 73.27\%, 73.64\%, and 73.77\%, respectively, confirming that each component contributes a non-trivial and complementary gain. Importantly, while the gain over the strongest prior method appears modest (0.75\%), the cumulative improvement over our own baseline is substantial: the baseline achieves 71.84\% mAP on SYSU-MM01 (all-search), HDCA increases it to 73.47\%, and GDL further pushes it to 74.95\%, totaling 3.11\% absolute improvement. On RegDB, V$\rightarrow$I mAP improves from 80.37\% to 80.96\% and I$\rightarrow$V from 77.66\% to 78.24\%, demonstrating that the gains stem from the proposed modules rather than from hyperparameter tuning.
\begin{figure}[!t]
	\centering
	
	
	\includegraphics[width=0.9\columnwidth]{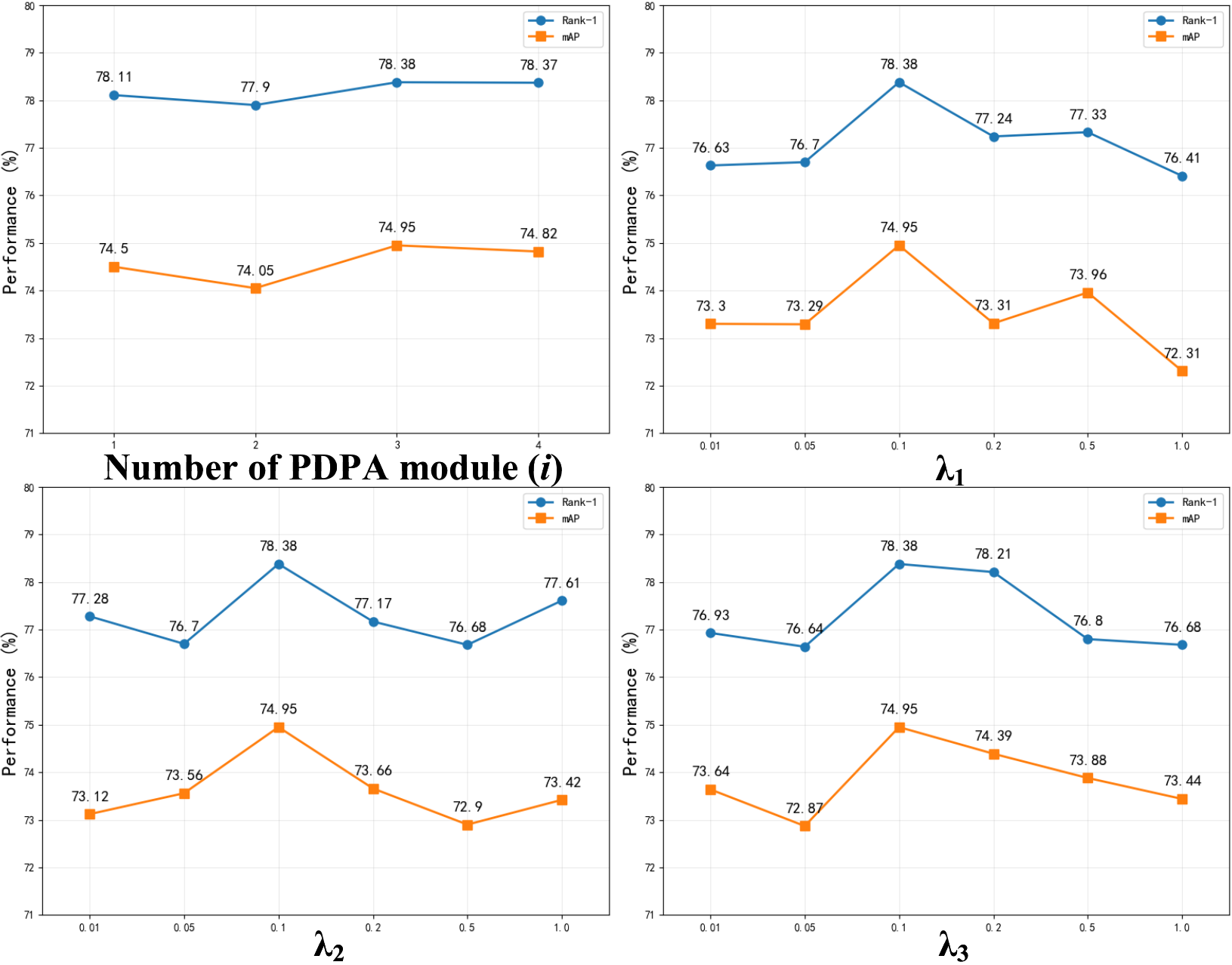}%
	
	\caption{Ablation study for the number of PDPA module ($i$) in Eq. \ref{eq2} and hyperparameter analysis of $\lambda_1$, $\lambda_2$ and $\lambda_3$ in Eq. \ref{eq9}.}
	\label{ab_combined}
\end{figure}
\begin{figure}[!t]
	\centering
    
	\includegraphics[width=1\columnwidth]{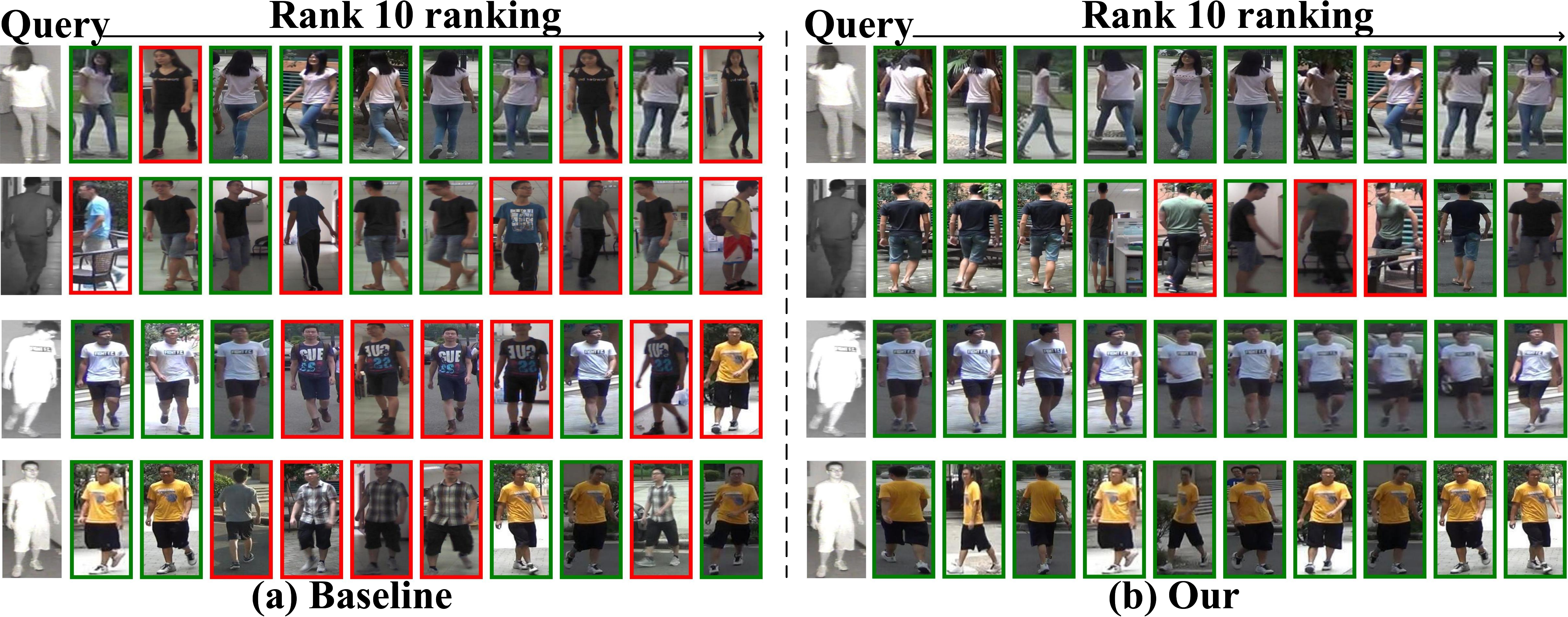}
	\caption{Visualization of R10 retrieval results for visible-infrared images on SYSU-MM01. \textcolor{red}{Red} for wrong and \textcolor{green}{green} for correct matches.}
	\label{vis_combined}
\end{figure}

\noindent\textbf{Hyperparameter Analysis.} 
To assess the impact of key hyperparameters in MDCRNet, we conduct systematic experiments on the SYSU-MM01 dataset under all-search mode. We evaluate the number of PDPA modules, namely $i$, and the weighting coefficients for $\mathcal{L}_{GD}$ components. As shown in Fig. \ref{ab_combined}, optimal performance is achieved with 3 PDPA modules and balanced weighting of $\lambda_1$=0.1, $\lambda_2$=0.1, $\lambda_3$=0.1 for the $\mathcal{L}_{GD}$ components. Analysis reveals that using 3 PDPA modules effectively captures multi-scale dependencies without introducing excessive computational overhead, while the uniform 0.1 weighting maintains optimal balance between intra-identity compactness and inter-identity separation across modalities. Sensitivity analysis further shows that $\lambda_1<0.08$ causes insufficient cross-modal alignment, while $\lambda_1>0.12$ over-compresses features and harms discriminability; $\lambda_2$ remains stable across a wide range; $\lambda_3>0.12$ leads to a sharp decline due to over-regularization that suppresses feature diversity. 

Notably, performance drops when only two PDPA modules are used. This validates our multi-scale design: incomplete hierarchies (short + medium only) suffer from feature imbalance and lack critical long-range semantic information, confirming that all three distance scales are necessary for comprehensive cross-modal feature extraction.

\noindent\textbf{Visualization Analysis.} 
As illustrated in Fig. \ref{vis_combined}, our MDCRNet demonstrates superior retrieval performance with significantly more correct matches compared to the baseline method.

\noindent\textbf{Stability Analysis.} 
To verify that our improvements are not artifacts of random seed variation, we run MDCRNet 5 times on SYSU-MM01 (all-search) with different random seeds. The mean mAP is 74.95\% with a standard deviation of $\pm 0.10\%$, confirming that the reported gains are consistent and statistically meaningful.

\section{Conclusion}
\label{sec:conclusion}

This paper proposes MDCRNet that effectively mitigates cross-modal discrepancies through hierarchical architectural design and discriminative metric learning. Our approach introduces the HLM with efficient CA and PDPA to capture multi-scale cross-modal semantic information across short, medium, and long-range spatial dependencies. Moreover, we develop the JDML that synergistically combines identity loss, QCT loss, and our novel GDL to enforce fine-grained feature discrimination across different modalities and identities. Extensive experiments on SYSU-MM01 and RegDB datasets demonstrate the superiority of our method, achieving 0.75\% and 1.06\% mAP gains over the strongest competing methods on SYSU-MM01 and RegDB respectively, as well as a cumulative 3.11\% mAP improvement over our own baseline on SYSU-MM01. In future work, we will explore incorporating skeletal information extraction and multi-granularity semantic guidance to further enhance cross-modal refinement capability.

\begin{credits}
\subsubsection{\ackname}
This work was supported in part by the National Natural Science Foundation of
China under Grants 62562058 and 62441213, the Key R\&D Program of Xinjiang
Uygur Autonomous Region under Grant 2022B01046, and the Sichuan Province
Joint Fund Project of China under Grant 25QYCX0103.

\subsubsection{\discintname}
The authors have no competing interests to declare that are relevant to the
content of this article.
\end{credits}
%
%
%
%
\bibliographystyle{splncs04}
\bibliography{references}
\end{document}